\pdfoutput=1

\documentclass[11pt]{article}

\usepackage[]{ACL2023}

\usepackage{times}
\usepackage{latexsym}

\usepackage[T1]{fontenc}

\usepackage[utf8]{inputenc}

\usepackage{microtype}

\usepackage{inconsolata}
\usepackage{booktabs, multirow} 
\usepackage{graphicx}

\title{ClinicalMamba: A Generative Clinical Language Model on Longitudinal Clinical Notes}

\author{
Zhichao Yang$^{1}$, \ Avijit Mitra$^{1}$, \ Sunjae Kwon$^{1}$,  \ Hong Yu$^{1, 2}$ \\
$^{1}$ College of Information and Computer Sciences, University of Massachusetts Amherst\\
$^{2}$ Department of Computer Science, University of Massachusetts Lowell\\
{\tt \{zhichaoyang,avijitmitra,sunjaekwon\}@umass.edu } {\tt hong\_yu@uml.edu } \\
}

\begin{document}
\maketitle
\begin{abstract}
The advancement of natural language processing (NLP) systems in healthcare hinges on language models' ability to interpret the intricate information contained within clinical notes. 
This process often requires integrating information from various time points in a patient's medical history. 
However, most earlier clinical language models were pretrained with a context length limited to roughly one clinical document. 
In this study, We introduce ClinicalMamba, a specialized version of the Mamba language model, pretrained on a vast corpus of longitudinal clinical notes to address the unique linguistic characteristics and information processing needs of the medical domain. 
ClinicalMamba models, with 130 million and 2.8 billion parameters, demonstrate superior performance in modeling clinical language across extended text lengths compared to Mamba and clinical Llama. 
With few-shot learning, ClinicalMamba achieves notable benchmarks in speed and performance, outperforming existing clinical language models and large language models like GPT-4 in longitudinal clinical tasks. 
\end{abstract}

\section{Introduction}

Clinical narratives, such as patient histories, consultation notes, and discharge summaries, contain detailed and complex information that extends over long text sequences \citep{Wu2019Deep}. To fully understand a patient's condition, treatments, and outcomes, NLP systems need to integrate information from various parts of these narratives, which often requires understanding the context provided in those long form text \citep{Blumenthal2010LaunchingH}. 

Understanding the sequence of health events is crucial for diagnoses, treatment plans, and patient monitoring \citep{wang2024recent, Yang2023TransformEHRTE, Eva2005What}. This often involves putting together information from different time points within a patient's health history \citep{gao2024units}. Long context enables NLP systems to perform temporal reasoning by tracking events over time longitudinally, which is essential for tasks like predicting disease progression or extracting medical relation \citep{Chen2023LanguageMA, jia-etal-2019-document,wiegreffe-etal-2019-clinical}.

It becomes imperative to design models for the need for processing longer texts \citep{Parmar2023LongBoXET,Tay2020LongRA}. 
Prior studies have introduced Mamba \citep{Gu2023MambaLS}, a selective state space model, 
that selects and compresses all necessary information into latent space from context, 
and achieves linear-time efficiency with context length. While these advancements have been primarily directed towards processing general domain text, the unique linguistic features of clinical narratives differ significantly from general domain \citep{Lehman2023DoWS}, motivating us to develop specialized Mamba models in the clinical domain. 

In this work, we build and publicly release ClinicalMamba - a Mamba model pretrained on longitudinal clinical notes. Furthermore, we demonstrate that ClinicalMamba outperforms multiple language models on longitudinal clinical NLP tasks. In particular, our contributions are as follows:

\begin{itemize}
\item We publicly release ClinicalMamba with 130m and 2.8b parameters trained on MIMIC-III \citep{Johnson2016MIMICIIIAF}.\footnote{\url{https://github.com/whaleloops/ClinicalMamba}} 

\item Through distributed training,  ClinicalMamba-2.8b model was pretrained in under 60 hours on 4 A100 GPUs and it is the first clinical autoregressive language model with a 16k maximum token length.

\item Through few-shot prompt-based finetuning, we demonstrate both ClinicalMamba outperforms original Mamba, GPT4, and other existing clinical long context language models on well-established long context clinical information extractions tasks: cohort selection for clinical trial and international classification of diseases (ICD) coding.
\end{itemize}

\begin{figure}[t]
	\centering
	\includegraphics[width=0.48\textwidth]{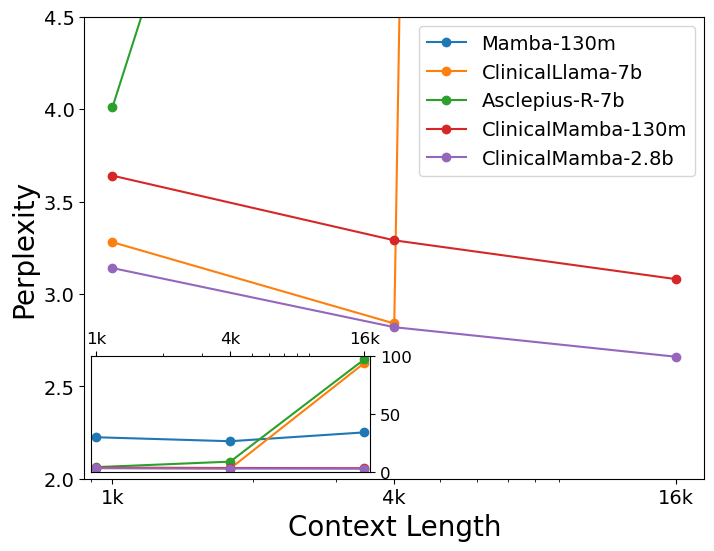}
	\caption{Perplexity of different generative language models on MIMIC-III when evaluated at various preceding context lengths (1k, 4k, and 16k tokens). The X-axis is in the log scale. The subfigure is a zoom-out plot with perplexity ranges 0-100. Experiment settings and detailed results are in section \ref{sec:results}.}
	\label{fig:perplexity}
\end{figure}

\section{Related Work}

\subsection{Pretraining clinical narratives}
The rapid expansion of the utilization of electronic health records (EHRs) into the healthcare landscape underscores an urgent need for a clinical language model \citep{Kang2019PretrainingTR}. Previous work, such as \citep{alsentzer-etal-2019-publicly} on Clinical BERT embeddings and \citep{Huang2019ClinicalBERTMC} with ClinicalBERT, adapted general-purpose language models to the clinical domain to enhance performance on clinical tasks. These models have been pivotal in demonstrating the effectiveness of adapting general-purpose NLP tools to the intricacies of clinical text. Similarly, the creation of GatorTron \citep{Yang2022ALL} scales up clinical language models to billions of parameters, while NYUTron \citep{Jiang2023HealthSL} harness billions of unstructured data found in EHRs. Both underscores the potential of domain-adapted language models to advance clinical NLP by improving performance across various tasks such as concept extraction and outcome prediction \citep{Yang2022ALL, Jiang2023HealthSL}.

To handle complex and nuanced tasks, recent studies investigated training generative models with prompt \citep{Kweon2023PubliclySC,Peng2023ASO,wang2023large,lu-etal-2022-clinicalt5,wang-sun-2022-promptehr}. These models not only excel in classification but also in generating clinically relevant text that can be indistinguishable from human-written notes.
Most previous methods focus on pretraining transformer models with a context window less than 2k tokens. However, we pretrained a selective state space model with a context window of 16k tokens, which includes more than 98\% of the visits in MIMIC-III.

\subsection{Clinical information extraction on long document}
Handling long texts in clinical NLP has always been challenging. Traditional methods of information extraction tackle this by marking specific locations within the sentence, but such labeling is not always available, and hiring annotators can be costly \citep{Fu2020ClinicalCE, kwon-etal-2022-medjex, deshpande2024localtweets}. Recent advancements in document information extraction involve pairing labels with documents. However, BERT struggles with processing these lengthy documents directly. 

To address this, prior research introduced \textit{Hierarchical-ClinicalRoberta}, which involves breaking down long documents into shorter segments of 512 tokens, applying ClinicalRoberta to each segment to obtain embeddings, and then using additional layers to leverage these embeddings for label classification \citep{huang-etal-2022-plm, Zhang2022HierarchicalBF}. However, this method combines information from each segment only at the final layer, which can hinder performance when training data is limited.

To mitigate this issue,  \textit{ClinicalLongformer} is designed to efficiently process longer context length by employing a self-attention mechanism across all layers, which is key to its proficiency in managing dense information exchanges within a specified contextual range \citep{Li2022ClinicalLongformerAC, Ji2023DomainspecificCP}. This mechanism, while powerful, is limited by its focus on a predetermined window of text, restricting its scope to what falls within this window. 

To overcome these limitations, the Mamba model emerges as a revolutionary approach. Mamba employs a selective state space model strategy to meticulously choose critical data for incorporation into its state \citep{Gu2023MambaLS}, thereby, enhancing its capability to manage information beyond the conventional self-attention window. In general domain language modeling, Mamba surpasses Transformers of equivalent size in task performance and speed.

\section{Methods}

\subsection{Pretraining}
We gather 82,178 hospital visits along with their deidentified free-text clinical notes (2,083,180) from 46,520 patients in MIMIC-III \citep{Johnson2016MIMICIIIAF}. Rather than breaking down the notes into chunks of 512 tokens to act as individual data instances, we aggregate all notes related to a visit longitudinally. The distribution of token counts per data instance is detailed in Table \ref{tab:data}. For information on our text pre-processing methods, please refer to section \ref{sec:textproc}.

Following previous works \citep{fei2019finetune, alsentzer-etal-2019-publicly}, we continue to pretrain Mamba using MIMIC-III clinical notes with the causal language modeling objective. 
This pretraining process utilizes 4 Nvidia A100-80GB GPUs. It's important to note that some of our downstream evaluation tasks utilize a small subset (6,049) of hospital visits from MIMIC-III, so we exclude them from the pretaining data. A comprehensive training recipe is available in section \ref{sec:training_recipe}.

\subsection{Prompt based fine-tuning}
We leverage the inherent capabilities of pre-trained language models by introducing a novel fine-tuning strategy that aligns with the specific demands of few-shot learning in clinical NLP. Recognizing the limitations of traditional fine-tuning methods when applied to clinical NLP tasks with limited labeled data, we adapted a prompt-based fine-tuning mechanism following previous works \citep{gao-etal-2021-making,yang-etal-2022-knowledge-injected,taylor2023clinical}. Specifically, we first identify a set of representative prompts that encapsulate key aspects of the clinical tasks, such as the patient's alcohol consumption. These prompts are then appended after each input clinical note and incorporated into the fine-tuning phase, where the language model learns to associate them with label tokens (Yes/No) based on a limited dataset. The generated label tokens are then mapped to label space. 

\begin{figure}[t]
	\centering
	\includegraphics[width=0.48\textwidth]{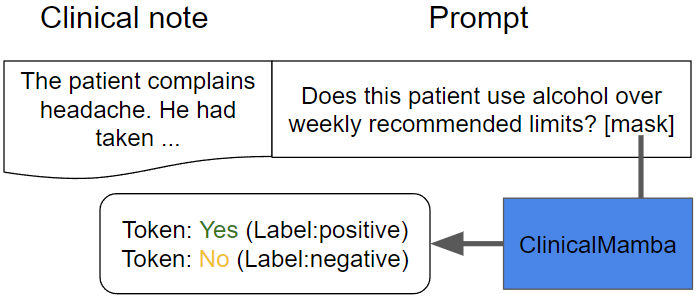}
	\caption{Illustration of Prompt-based fine-tuning.}
	\label{fig:prompt}
\end{figure}

As shown in Figure \ref{fig:prompt}, we transfer the downstream information extraction task into a pretraining-like task - label token generation.

\subsection{Fine-tuning tasks}

Cohort selection for clinical trial addresses the challenge of interpreting unstructured clinical narratives to streamline the patients selection process. It aims to classify patients based on whether they meet 13 specific eligibility criteria, such as the usage of aspirin to prevent myocardial infarction, excessive alcohol consumption, and HbA1c values between 6.5 and 9.5\%, among others. The input contains multiple clinical notes with a total length of 4924 tokens on average. This dataset was released as part of n2c2 challenge (track 1) in 2018.


ICD coding interprets complex clinical narratives, translating them into standardized codes that facilitate accurate billing, statistical analysis, and healthcare management. It aims to extract patient's disease and procedure codes from clinical text. We followed general instructions from \citet{mullenbach-etal-2018-explainable} in building this task, but instead of using a single discharge summary as input, we used all previous discharge summaries and assigned ICD code descriptions from previous visits. We further filtered 50 infrequent codes as Code-rare and 50 frequent codes as Code-common following \citet{yang-etal-2022-knowledge-injected}. The average length is 4,223 and 7,062 tokens respectively. Detailed dataset statistics are shown in Table \ref{tab:data}.

For each task, we report the micro precision, recall, F1 scores, and the receiver operating characteristic/area under the curve (ROCAUC) on the test dataset.

\section{Results \& Discussions}
\label{sec:results}

\begin{table}[ht]
\centering
\begin{tabular}{lrrrrr}\toprule
Model &Prec &Recall &F1 &AUC \\\midrule
\textbf{C}Llama2 &70.0 &79.1 &77.7 &84.3 \\
Hi-\textbf{C}Roberta &72.4 &82.6 &79.2 &88.1 \\
\textbf{C}Longformer &69.7 &78.6 &76.1 &83.5 \\
GPT-4 &88.1 &79.9 &84.8 &- \\
Mamba-130m &75.4 &80.2 &77.7 &85.7 \\\midrule
\textbf{C}Mamba-130m &79.0 &86.2 &82.2 &91.8 \\
\textbf{C}Mamba-2.8b &\textbf{88.6} &\textbf{89.5} &\textbf{88.8} &\textbf{95.7} \\
\bottomrule
\end{tabular}
\caption{Results on cohort selection task, where \textbf{C} is model pretrained in clinical domain.}
\label{tab:result_cohort}
\end{table}

\begin{table*}[ht]
\centering
\begin{tabular}{lrrrrrrrrr}\toprule
\multirow{2}{*}{Model} &\multicolumn{4}{c}{Code-rare} &\multicolumn{4}{c}{Code-common} \\\cmidrule{2-9}
&Prec &Recall &F1 &AUC &Prec &Recall &F1 &AUC \\\midrule
MultiResCNN &20.34 &2.07 &5.19 &47.2 &70.5 &60.78 &66.24 &92.04 \\
Hi-\textbf{C}Roberta &46.19 &10.96 &16.74 &77.11 &73.76 &65.01 &69.23 &93.14 \\
\textbf{C}Longformer &50.27 &17.81 &28.69 &80.52 &\textbf{78.42} &64.97 &71.14 &94.24 \\
GPT-4 &30.91 &36.12 &33.29 &- &72.48 &62.28 &68.19 &- \\
Mamba-130m &57.75 &28.08 &37.79 &84.8 &73.71 &62.87 &68.94 &92.75 \\\midrule
\textbf{C}Mamba-130m &70.97 &30.14 &42.31 &91.08 &76.82 &68.03 &\textbf{74.34} &94.23 \\
\textbf{C}Mamba-2.8b &\textbf{75.28} &\textbf{45.89} &\textbf{56.51} &\textbf{92.75} &75.53 &\textbf{72.12} &73.64 &\textbf{94.54} \\
\bottomrule
\end{tabular}
\caption{Results on ICD coding task, where \textbf{C} indicates model pretrained in clinical domain.}
\label{tab:result_icd}
\end{table*}

In this section, we will first compare the model's language modeling ability on MIMIC-III clinical notes.
We will then describe the evaluation on different clinical information extraction tasks. 
Finally, we will describe trade-offs between language modeling abilities (perplexity) and inference speed (throughput) for several generative models on MIMIC-III notes. Baseline models used are detailed in section \ref{sec:baselines}.

ClinicalMamba stands as the sole model capable of handling clinical notes of up to 16k tokens. As demonstrated in Figure \ref{fig:perplexity}, the perplexity for ClinicalMamba-2.8b decreased from 3.11 to 2.61 as the context length expanded from 1k to 16k tokens during inference. This is in contrast to the performance of prior clinical autoregressive language models, where perplexity levels rose with increased context lengths. For instance, with ClinicalLlama-7b, perplexity escalated from 2.82 to 94.02 as the context length grew from 4k to 16k. This limitation arises because these models were trained on contexts not exceeding 4k, impairing their accuracy for next token prediction when given previous contexts beyond 4k.

In the domain of extracting information from longitudinal clinical records, ClinicalMamba demonstrates superior performance compared to Mamba.
ClinicalMamba achieved ROCAUC scores of 91.8, 42.3, and 94.2 on Cohort selection, Code-rare, and Code-common, while Mamba obtained ROCAUC scores of 85.7, 37.8, and 92.8 respectively. 
ClinicalMamba also outperformed previous long-range clinical language models with similar number of parameters. ClinicalMamba significantly outperformed Hierachical-ClinicalRoberta and ClinicalLongformer by relatively 52.7\% and 19.1\% on ROCAUC respectively. This is particularly notable in the Code-rare task with limited training data (5 shots), where ClinicalMamba attained an AUC of 91.1, compared to 77.1 of Hierarchical-ClinicalRoberta and 80.5 of ClinicalLongformer.
Surprisingly, ClinicalMamba-2.8b also outperformed zero-shot GPT-4, achieving F1 scores of 88.8, 56.6, and 73.6 on Cohort selection, Code-rare, and Code-common tasks, whereas GPT-4 obtained a F1 score of 84.8, 33.3, and 68.2 respectively.

ClinicalMamba also offers a great tradeoff between language modeling abilities and inference speed. On one side, small language models' perplexity is limited. On the other side, the remarkable perplexity reduction delivered by large language models comes at a steep increase in computational cost. As shown in Table \ref{tab:tradeoff}, the perplexity of ClinicalMamba-2.8b (2.61) is comparable to that of ClinicalLlama-7b (2.82) trained with the same computation budget. Moreover, the inference speeds of ClinicalMamba-2.8b and ClinicalMamba-130m are 3 to 30 times faster than that of ClinicalLlama7b.

\section{Conclusion}
In this study, we developed and released Mamba models pretrained on a large collection of clinical notes. Our findings demonstrate the superior performance of our ClinicalMamba in extracting information from long text documents compared to other models. We strongly believe that clinical NLP researchers can benefit from such long-context generative language models that alleviates the need of a substantial computational power, without any performance trade-off. Building on the groundwork laid by this study, future endeavors can further refine and expand the capabilities of ClinicalMamba, thereby enhancing the effectiveness of clinical data processing across diverse medical fields.

\section*{Limitations}
This work has several notable limitations. First, we do not experiment with more recent parameter-efficient fine-tuning strategies such as soft prompting \citep{lester-etal-2021-power} and Low-Rank Adaptation (LoRa) \citep{Hu2021LoRALA}. This potentially undermined ClinicalMamba on downstream tasks. Second, our adaptation of the Mamba framework was restricted solely to textual data documented during visits. EHRs are rich with multifaceted information, including but not limited to radiology images taken at different times and Electrocardiogram waveforms that span various periods. Future research could develop a multimodal Mamba framework to leverage all other modalities. Third, the MIMIC-III dataset, which serves as the foundation of our study, only includes notes from the intensive care unit of a single hospital within the United States. This limits the generalizability of our findings, as care practices vary significantly across different institutions and countries. We did not pretrain on MIMIC-IV because it only has a limited number of notes (and also limited type: discharge summary and radiology report) per visit. Lastly, the linguistic scope of the MIMIC dataset is limited to English, which presents a barrier to understanding and applying our findings in non-English speaking contexts. Addressing these limitations could substantially broaden the applicability and relevance of our work in future endeavors.

\section*{Ethics Statement}
In this research, we gained authorized access to the MIMIC and N2C2 dataset and used de-identified clinical notes following their license agreement and HIPAA regulations. When language models are trained on extensive clinical text, they can inherit biases within the data. For instance, they might prefer inquiries concerning smoking habits or link specific medical conditions to certain demographic groups. These biases could be mitigated by enhancing model alignment with each patient’s background.

\bibliography{anthology,custom}
\bibliographystyle{acl_natbib}

\newpage
\appendix
\renewcommand\thefigure{\thesection.\arabic{figure}}  
\renewcommand\thetable{\thesection.\arabic{table}}

\section{Appendix}
\setcounter{figure}{0}
\setcounter{table}{0}

\subsection{Text preprocessing}
\label{sec:textproc}

\begin{table*}[ht]
\centering
\begin{tabular}{lrrrrr}\toprule
& &Cohort selection  &Code-rare &Code-common \\\midrule
shots &mean &89 &5 &918 \\\midrule
\multirow{4}{*}{tokens} &mean &4924  &4223 &7062 \\
&median &4632  &3236 &5177 \\
&99\% &10781  &14345 &13356 \\
&max &13989 &18480 &14773 \\
\bottomrule
\end{tabular}
\caption{Number of instances per label (shots) and number of tokens per input.}
\label{tab:data}
\end{table*}

We followed \citet{Huang2019ClinicalBERTMC} to format notes during text preprocessing. But We did not convert text to lowercase because Mamba tokenizer is able to process both upper and lower cases. For notes on each patient's hospital visit, we sorted notes by their charted date and concatenated notes into one string. We used string "- - \{NoteType\} note  - -" to separate the notes. Table \ref{tab:mimiciii} shows comprehensive values of \{NoteType\}.

\begin{figure}[t]
	\centering
	\includegraphics[width=0.48\textwidth]{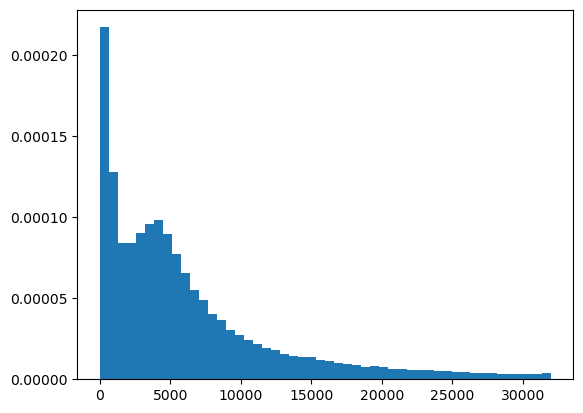}
	\caption{Long tail distribution of number of tokens per each visit. Y-axis is the density (sum to 1.0).}
	\label{fig:lendist}
\end{figure}

For pretraining data, we truncate notes with more than 16k tokens, however, this is only less than 2\%, a length distribution is provided in Figure \ref{fig:lendist}. We exclude a small subset (6,049) of hospital visits due to the evaluation of MIMIC ICD coding and MIMIC hospital readmission prediction, The visit ids (hadm\_id) are documented in the github.

\begin{table*}[ht]
\centering
\begin{tabular}{lrrrr}\toprule
Category &Count &\% &Len \\\midrule
Nursing &506,528 &73 &241 \\
Radiology &338,834 &83.3 &449 \\
ECG &123,042 &61.3 &43 \\
Physician &92,426 &18.2 &1369 \\
Discharge summary &47,572 &96.7 &2195 \\
Echo &34,064 &45.8 &464 \\
Respiratory &32,798 &8.1 &205 \\
Nutrition &7,971 &6.4 &602 \\
General &7,710 &6.4 &290 \\
Rehab Services &5,321 &4.6 &622 \\
Social Work &2,294 &2.8 &446 \\
Case Management &939 &1.3 &260 \\
Pharmacy &97 &0.1 &512 \\
Consult &78 &0.1 &1206 \\
\bottomrule
\end{tabular}
\caption{Statistic of note events documented in MIMIC-III dataset. Each column represents a) the number of notes, b) proportion of visits, c) average number of words for each note type.}
\label{tab:mimiciii}
\end{table*}

\subsection{Pretraining recipe}
\label{sec:training_recipe}
ClinicalMamba-2.8b is a selective state space model designed using replication of the Mamba architecture \citep{Gu2023MambaLS}. ClinicalMamba refers to the class of models, while 2.8b represents the number of parameters of this particular pretrained model. We also pretrained ClinicalMamba-130m using pretraining data from the previous section. The specific values of hyperparameters are shown in Table \ref{tab:hyperpeameter}. These models were trained for 763 million English tokens over 7000 steps (3 epochs) \citep{Muennighoff2023ScalingDL}. It was trained as an autoregressive language model, using cross-entropy loss \citep{NEURIPS2020_1457c0d6}. For learning rate scheduling, we followed Mamba and chose linear learning rate warmup with cosine decay to $1e-5$. We found this important setting to avoid loss overflow. It took under 60 hours to pretrain ClinicalMamba-2.8b in on 4 Nvidia Tesla A100-80GB GPUs.

\begin{table*}[ht]
\centering
\begin{tabular}{lrr}\toprule
Hyperparameter &Value \\\midrule
num param &130m/2.8b \\
num layer &24/64 \\
dim model &768/2560 \\
context len &16k \\
num vocab &50277 \\
position emb &None \\
optimizer &Adam \\
beta1 &0.9 \\
beta2 &0.95 \\
epsilon &1e-5 \\
batch size &32 \\
weight decay &0.1 \\
gradient clipping &1.0 \\
peak learning rate &1e-3/6e-4 \\
\bottomrule
\end{tabular}
\caption{Hyperparameters used to train ClinicalMamba.}
\label{tab:hyperpeameter}
\end{table*}

\subsection{Baselines}
\label{sec:baselines}

\noindent \textbf{GPT-4} is a large language model designed to understand and generate human-like text based on the input it receives. We applied zero-shot prompting to each downstream task, using ACAN and original prompt introduced in \citet{Wornow2024ZeroShotCT}.
GPT-4 (version 2023-12-01-preview) was accessed securely through the Azure OpenAI API. We set the sampling temperature for decoding to 0.1.

\noindent \textbf{Asclepius-R} \citep{Kweon2023PubliclySC} is a clinical generative language model trained on MIMIC-III discharge summaries and corresponding instruction-answer pairs. It has 7 billion parameters with a maximum input of 4096 tokens. 

\noindent \textbf{ClinicalLlama2} (CLlama2) is similar to Asclepius-R, but it was trained on all types of MIMIC-III note with the same computation budget as ClinicalMamba-2.8b (60 hours on 4 A100). It has 7 billion parameters with 4096 max context length.

\noindent \textbf{ClinicalLongformer} (CLongformer) \citep{Li2022ClinicalLongformerAC} is a clinical knowledge enriched version of Longformer that was further pretrained using MIMIC-III clinical notes. It has 149 million parameters with a maximum input of 4096 tokens. We only used local attention and does not apply global attention for computation efficiency. 

\noindent \textbf{Hierachical ClinicalRoberta} (Hi-CRoberta) \citep{huang-etal-2022-plm}, utilizes multiple embedding from clinical Roberta \citep{lewis-etal-2020-pretrained}. It first segment clinical notes into chunks of 512 tokens to obtain their embeddings, embeddings are then pooled by concatenation and finally a linear classification head during downstream task. It has 110 million parameters with a max of 16384 tokens.

\noindent \textbf{MultiResCNN} \citep{Li2020ICDCF} encode free text with Multi-Filter ResidualCNN, and applied label code attention mechanism to enable each ICD code to attend different parts of the document.

\begin{table*}[ht]
\centering
\begin{tabular}{lrrrr}\toprule
&Context Length &Perplexity &Throughput \\\midrule
Pythia-130m &1k &\textbf{34.86} &12042 \\
&4k &69.79 &14859 \\
&16k &566.42 &15713 \\
Mamba-130m &1k &29.90 &24695 \\
&4k &\textbf{26.45} &34539 \\
&16k &34.11 &37009 \\
ClinicalMamba-130m &1k &3.64 &24709 \\
&4k &3.29 &33349 \\
&16k &\textbf{3.08} &35486 \\
ClinicalLlama2-7b &1k &3.28 &1005 \\
&4k &\textbf{2.82} &1013 \\
&16k &94.02 &951 \\
Asclepius-R-7b &1k &\textbf{4.01} &1066 \\
&4k &8.79 &1064 \\
&16k &97.37 &995 \\
ClinicalMamba-2.8b &1k &3.11 &2932 \\
&4k &2.84 &3007 \\
&16k &\textbf{2.61} &3027 \\
\bottomrule
\end{tabular}
\caption{Trade-off between language modeling abilities (perplexity) and inference throughput (Tokens/s) for a number of models on MIMIC-III clinical notes. Among 3 variations of 1k, 4k, and 16k context length, the variation with best perplexity is bold for each model.}
\label{tab:tradeoff}
\end{table*}

\end{document}